# Multi-UAV Formation Cooperative Obstacle Avoidance and Adaptive Shape Deformation Control in Complex Environments Based on BI-APF-RRT and Affine Transformation

## Yiliang Wu [1], Weican Chen [1], Yendo Hu [1]*,

[1]College of Marine Information Engineering, Jimei University, Xiamen, China

*Corresponding author:201861000005@jmu.edu.cn

**Abstract.** Aiming at the problem that obstacle avoidance flexibility and formation integrity are difficult to coexist in multi-UAV formation motion in complex obstacle environments , and that the traditional artificial potential field (APF) method easily falls into local optima, a cooperative obstacle avoidance algorithm for multi-UAV formations integrating BI-APF-RRT and affine transformation is proposed. First, abandoning the traditional APF centroid path planning method , a goal-biased Bidirectional Artificial Potential Field method RRT (BI-APF-RRT) algorithm is adopted to conduct global collision-free path planning for the centroid of the leader formation. By introducing an improved artificial potential field and cubic B-spline interpolation, the smoothness and rapid convergence of the global path are ensured. Secondly, using the generated global path as the guiding trajectory for the formation's centroid , combined with an affine transformation matrix (including non-uniform scaling and rotation) , the formation can adaptively deform based on the distance to obstacles while moving along the optimal path. Finally, the followers track the leaders through a distributed control law , enabling the entire formation to safely cross complex obstacle areas without disassembling.

## 1. Introduction

Propelled by recent advances in autonomous control technologies, multi-UAV formations and unmanned aerial vehicle (UAV) swarms have been extensively deployed across diverse sectors, including agriculture, military operations, logistics, and industrial manufacturing [1-4]. Compared to individual UAVs, multi-agent cooperative systems exhibit overwhelming superiority in executing intricate real-world missions. Nevertheless, as operational environments become increasingly complex and unpredictable, navigating safely through unknown static or dynamic obstacles while preserving the coordinated topological structure of the formation has emerged as a paramount challenge and a focal research area in multi-agent control [5].

Current strategies for formation obstacle avoidance are primarily categorized into local autonomous evasion and formation switching. In the local autonomous approach, the formation temporarily disbands upon encountering threats, allowing each UAV to navigate independently before regrouping [6-7]. This decoupled paradigm severely disrupts the topological coherence of the swarm and incurs prohibitive time costs for structural reconfiguration, thereby degrading mission continuity. Alternatively, formation switching relies on transitioning among a predefined set of rigid configurations to bypass obstacles [8-9]. However, this method inherently lacks the flexibility required to adapt to unknown, sudden, or irregularly narrow terrains.

To reconcile the inherent conflict between maneuverability and spatial coherence during obstacle avoidance, researchers have introduced affine transformation theory into formation control [10]. By incorporating rotation, translation, non-uniform scaling, and shearing, affine transformations transcend the limitations of traditional conformal mappings (which only permit uniform scaling), enabling continuous and adaptive elastic deformations. While this approach has demonstrated tremendous potential in UAV and AUV swarms [11-12], these systems conventionally rely on the Artificial Potential Field (APF) method to generate global guiding trajectories for the formation centroid or leaders [13]. A critical flaw of traditional APF is its susceptibility to local minima; when navigating U-shaped obstacles or intricate narrow corridors, the cancellation between the attractive force of the target and the repulsive forces of obstacles frequently traps the centroid, causing the entire formation to stall.

Concurrently, sampling-based global planning algorithms, prominently represented by the Rapidly-exploring Random Tree (RRT), excel at resolving high-dimensional constraints but suffer from inherent randomness, leading to blind searching and excessive redundant nodes [14]. Although subsequent variants have integrated potential field guidance (APF-RRT) [15] and bidirectional search mechanisms (BI-RRT) [16] to accelerate convergence—and advanced algorithms like BI-APF-RRT* have achieved asymptotic optimality even in complex narrow spaces [17]—the rigorous node rewiring required to guarantee this optimality imposes a substantial online computational burden. Such high latency renders them inadequate for the stringent real-time demands of dynamic multi- UAVs formation obstacle avoidance.

To overcome the dilemma between spatial flexibility and structural integrity, and to bypass the local minima trap of traditional APF, this paper innovatively proposes a cooperative formation control framework that embeds the highly efficient, globally searchable BI-APF-RRT algorithm into affine transformation models.

Firstly, departing from conventional APF-based centroid planning, we utilize a goal-biased Bidirectional Artificial Potential Field Rapidly-exploring Random Tree (BI-APF-RRT) [18] to delineate a global collision-free path for the leader formation's centroid. This planner employs a goal-biasing strategy to direct the alternating expansion of dual trees, significantly boosting convergence speed, while integrating a stochastic perturbation mechanism into the potential field to guarantee an escape from local minima deadlocks. Secondly, a cubic B-spline interpolation algorithm is applied to smooth the discrete path, ensuring it satisfies the kinematic constraints of UAV. Ultimately, guided by this optimized global trajectory, the formation leverages the non-uniform scaling properties of the affine transformation matrix to adaptively adjust its scaling ratio and rotation angle in real time based on proximity to local obstacles. This synergistic paradigm seamlessly integrates macro-level optimal path tracking with micro-level elastic collision avoidance, empowering the multi-UAV system to efficiently navigate highly constrained environments without severing its physical connectivity.

## 2. System Model and Problem Description

This section establishes the environmental space model, the communication topology, and the UAV kinematic model for a multi-UAV formation operating in a complex two-dimensional (2D) environment. Building upon these models, the cooperative obstacle avoidance control objectives based on the BI-APF-RRT algorithm and affine transformation are explicitly formulated.

### 2.1 Environmental Modeling and Problem Formulation

To formulate the path planning problem for the multi-UAV formation, the system environment is defined within a 2D workspace. Let $F$ denote the entire search space. Assuming the presence of various static or dynamic obstacles within this space, the region occupied by all obstacles is defined as $F_{obs}$. Consequently, the obstacle-free space where the multi-UAV formation can safely navigate is given by $F_{free} = F \setminus F_{obs}$.

Let $F_{start}$ and $F_{goal}$ denote the initial and target positions, respectively, of the centroid of the leader UAVs in the formation. The path planning problem, denoted as $\varpi\left(F, F_{start}, F_{goal}\right)$, aims to find a continuous, collision-free trajectory within $F_{free}$. Specifically, there exists a continuous function $\sigma: [0,1] \to F_{free}$ such that if $\sigma(0) = F_{start}$ and $\sigma(1) = F_{goal}$, then $\sigma$ represents a valid global, collision-free guiding trajectory from the starting point to the destination.

### 2.2 Communication Topology of the Multi-UAV Formation

Consider a multi-UAV formation system comprising $N$ mobile UAVs. The information exchange among the UAVs is modeled by a directed graph $G = (V, \epsilon)$, where $V = 1,2, \dots, N$ is the set of nodes (representing the UAVs), and $\epsilon \in V \times V$ is the set of directed edges. A directed edge $(j, i) \in \epsilon$ indicates that UAV $i$ can receive the state information from UAV $j$.This paper adopts a Leader-Follower architecture. The first $N_l$ UAVs are designated as leaders, forming the node subset $V_l = 1,2, \dots, N_l$; the remaining $N_f = N - N_l$ UAVs act as followers, comprising the node subset $V_f = V \setminus V_l$, as illustrated in Figure 1.

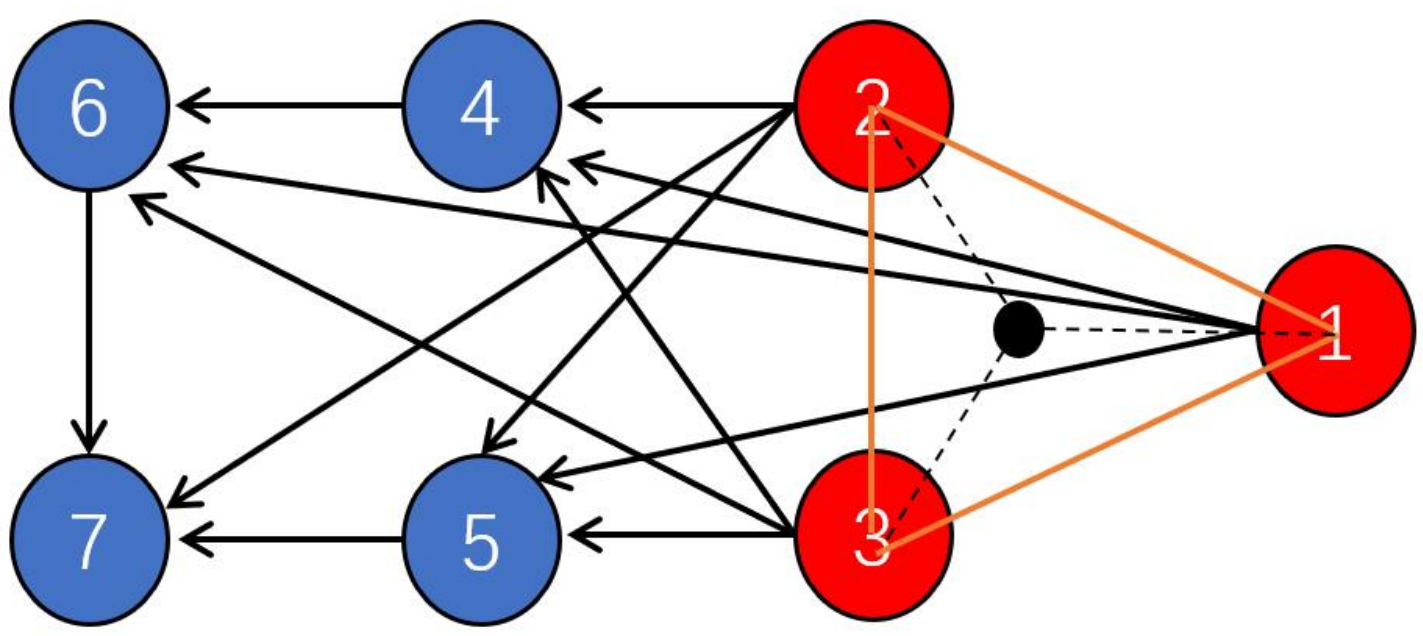

Fig. 1 This diagram shows the interaction topology of a drone formation, with red representing the leader, blue representing the followers, and black representing the centroid of the leader.

The interaction topology of the entire system can be characterized by a Signed Laplacian matrix, $L^s$. Based on the division of leaders and followers, the matrix $L^s$ can be partitioned into block matrices as follows:

$$L^s = \begin{bmatrix} 0^{N_l \times N_l} & 0^{N_l \times N_f} \\ L_{fl}^s & L_{ff}^s \end{bmatrix} \tag{1}$$

where $L_{ff}^s$ denotes the interaction topology among the follower UAVs, and $L_{fl}^s$ represents the interaction topology from the leaders to the followers.

### 2.3 UAV Kinematics and Affine Transformation Fundamentals

Let $p_i \in \mathbb{R}^2$ denote the position coordinates of the $i$-th UAV. For the $N_f$ followers within the formation, their kinematic behavior is modeled as a double-integrator system:

$$\begin{cases} \dot{p}_i = v_i \\ \dot{v}_i = u_i \end{cases} \tag{2}$$

where $v_i$ represents the velocity of the follower, and $u_i$ is the control input to be designed for the follower.

The essence of formation control lies in preserving relative position constraints. Suppose the nominal target configuration (Nominal Formation) of the team is defined by a set of vectors $g^* = [g_1^*, g_2^*, \dots, g_N^*]$. According to affine transformation theory, the desired positions of the actual formation can be derived by applying a linear transformation (encompassing rotation and non-uniform scaling) and a translation to the nominal configuration, satisfying:

$$p_i = Ag^* + b \tag{3}$$

where $A$ is a non-singular matrix serving as the affine transformation matrix (composed of a rotation matrix $A_{rot}$ and a scaling matrix $A_{sc}$), and $b$ is the translation vector. This transformation overcomes the constraints inherent in traditional rigid formations or similarity transformations (which permit only uniform scaling). It empowers the formation to dynamically compress or elongate along specific directions, thereby granting it the flexibility to traverse complex obstacles.

## 3. Core Algorithm Design

Traditional multi-UAV formation obstacle avoidance strategies based on affine transformations typically apply the standard Artificial Potential Field (APF) directly to plan the guiding trajectory for the leaders or the formation centroid. However, in complex physical environments (e.g., U-shaped obstacles or dense, narrow corridors), the attractive and repulsive forces of the traditional APF are highly prone to mutual cancellation in local regions. This causes the centroid to become trapped in local minima, subsequently leading to the stagnation of the entire formation.

To fundamentally overcome this bottleneck, this paper innovatively proposes the deep integration of the goal-biased Bidirectional Artificial Potential Field Rapidly-exploring Random Tree (BI-APF-RRT) algorithm into the "path-deformation" mapping mechanism. Rather than relying on purely local APF guidance, the proposed algorithm utilizes BI-APF-RRT to pre-plan a globally optimal, collision-free guiding path for the formation centroid. Subsequently, it employs an affine transformation matrix to drive the system to achieve adaptive deformation. This empowers the multi-UAV system with highly efficient obstacle avoidance capabilities in complex, constrained spaces while strictly maintaining the structural integrity of the formation topology.

### 3.1 Global Smooth Path Planning for Centroid Based on BI-APF-RRT

This section treats the centroid of the leader formation, $p_{cm}$, as a virtual exploration node, employing the BI-APF-RRT algorithm to generate a globally optimal path for it within the obstacle-free space. The innovative advantages of this algorithm are manifested in the following three aspects: High-Quality Sampling Strategy via Goal Biasing: To address the slow convergence caused by the blind global sampling of traditional RRT algorithms, a goal-biasing strategy is introduced. During the generation of random sampling points, a goal-bias threshold $m$ is established to directly select the target point $q_{goal}$ as the growth direction of the new node with a certain probability. Specifically, the random sampling node $q_{rand}$ is generated according to the following probability model:

$$q_{rand} = \begin{cases} random_state, & if\ rand() > m \\ q_{goal}, & otherwise \end{cases}$$

This alternately growing dual-tree structure effectively enhances sampling quality, substantially reducing the number of iterations and convergence time.

Potential Field Guidance with Local Minima Escape Mechanism: During the expansion of the random trees, an improved APF method is incorporated. The target point exerts an attractive force $F_{att}$, and obstacles exert a repulsive force $F_{rep}$, utilizing the resultant force vector to guide the growth of the tree branches. To completely eradicate the local minima trap, a distance-aware perturbation escape mechanism is established: when the Euclidean distance $\rho(q_{rand}, q_{obs})$ between the randomly expanded point and the nearest obstacle is less than a predefined safety threshold $\epsilon$, a random directional perturbation force $F_{random}$ is injected into the resultant force. The improved resultant force $F_{total}$ that guides the tree expansion is mathematically formulated as:

$$F_{total} = \begin{cases} \sum F_{rep} + \sum F_{att} & \epsilon > \rho(q_{rand}, q_{obs}) \\ \sum F_{rep} + \sum F_{att} + F_{random} & \epsilon < \rho(q_{rand}, q_{obs}) \end{cases} \quad (4)$$

where $\rho(q_{rand}, q_{obs})$ denotes the Euclidean distance between the sampled node and the nearest obstacle, and $\epsilon$ is the predefined safety threshold, attractive and repulsive forces are cited from the classic APF model. This stochastic perturbation forcibly breaks the equilibrium between the attractive and repulsive forces, enabling the centroid to successfully escape local dead ends, Figure 3 shows the algorithm process of APF-RRT, and the corresponding pseudocode is shown in Figure 2 below:

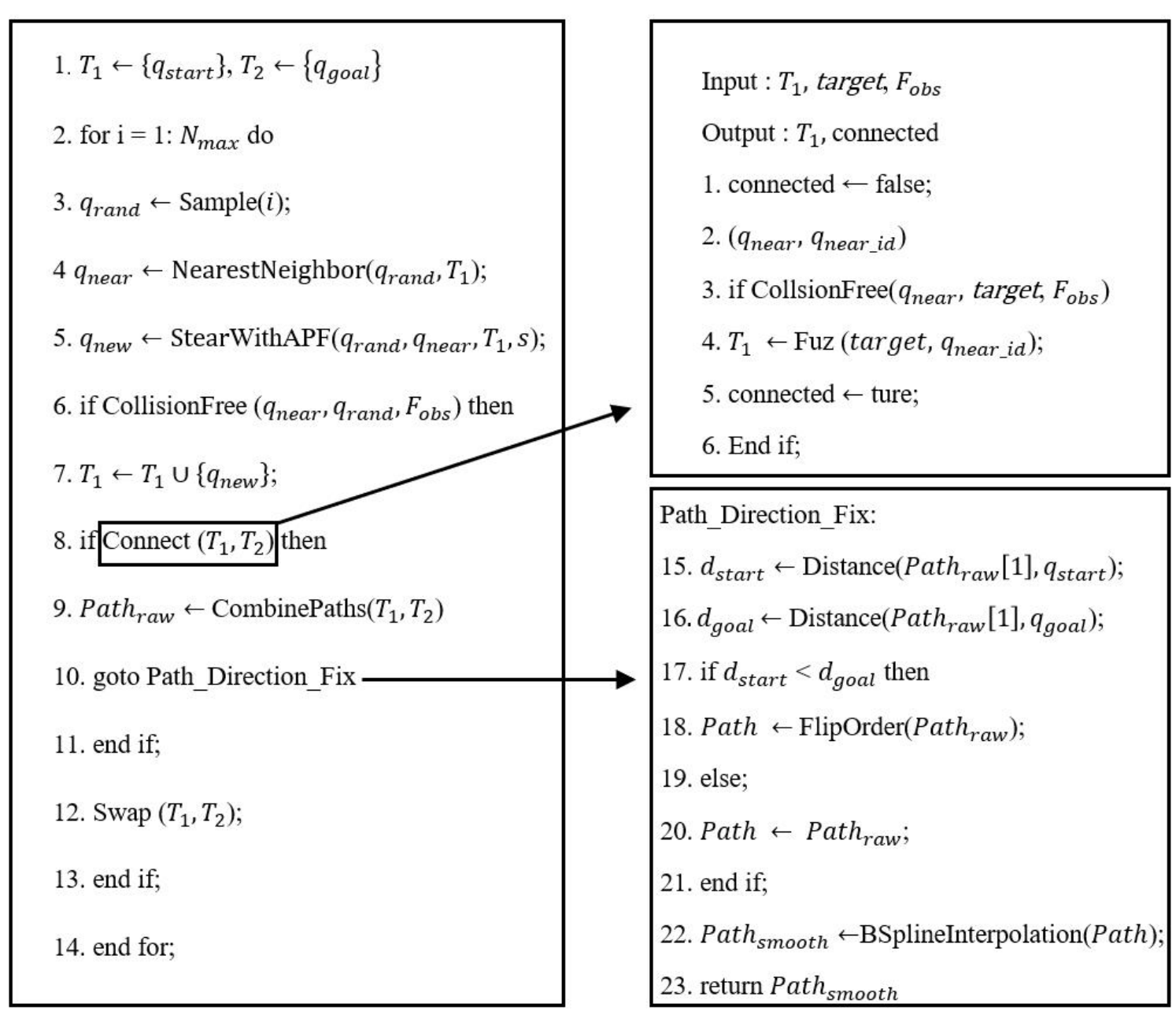

Fig. 2 BI-APF-RRT pseudocode. The left side is the main flow algorithm, with inputs being the start point, end point, step size $s$, and maximum number of iterations $N_{max}$, and the output being the ordered path Path from the start point to the end point. The right side shows the connect algorithm and the path direction correction algorithm.

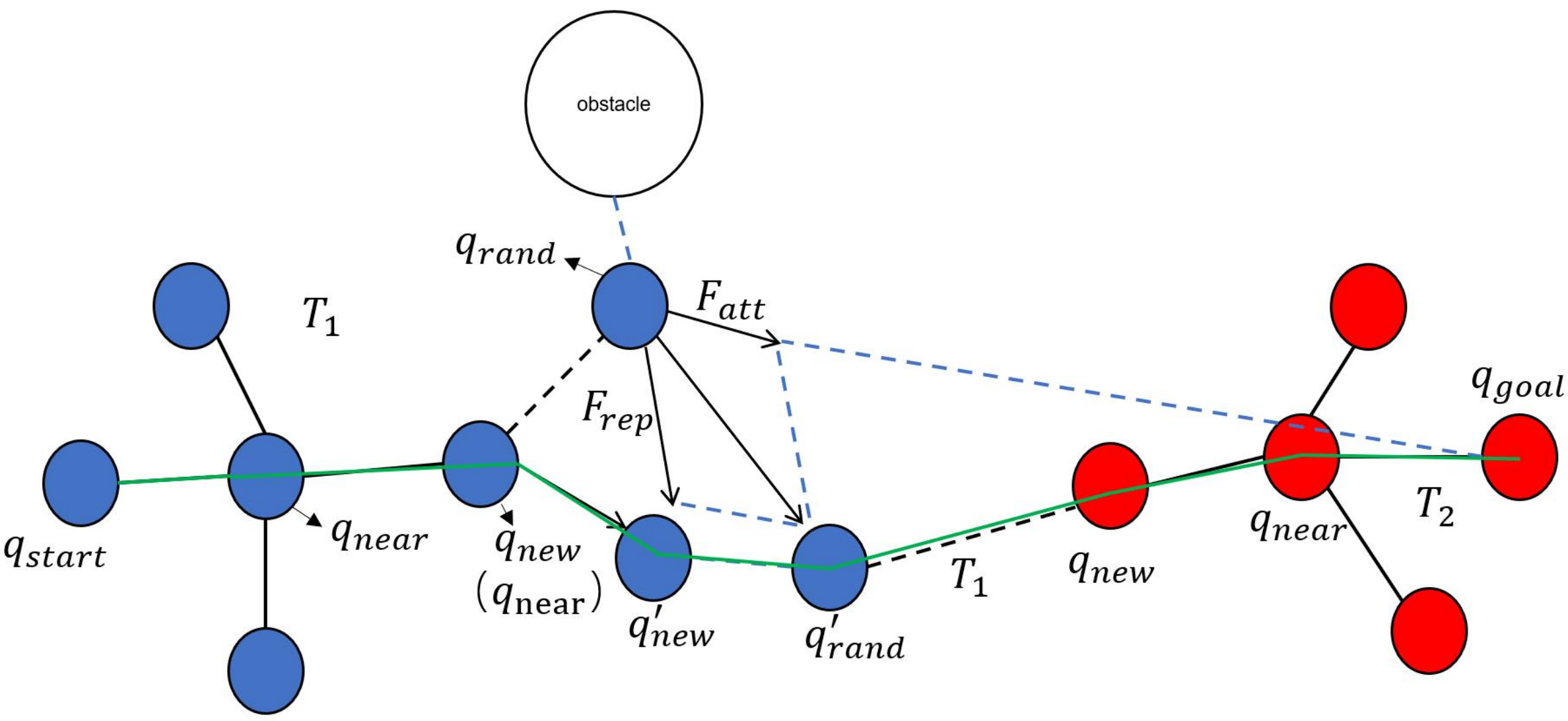

Fig. 3 BI-APF-RRT Node Expansion and Path Formation Process

B-Spline Smoothing Satisfying Kinematic Constraints: Since the original path generated by random trees consists of piecewise linear segments, it fails to satisfy the kinematic characteristics of physical UAVs. Therefore, a cubic B-spline interpolation method is adopted to smooth the path nodes, Figure 4 illustrates the optimization of path smoothing using cubic B-spline curves. The optimized discrete collision-free path points are stored as the set $P_{TAPE}$, providing a global reference with continuous curvature for the subsequent formation deformation.

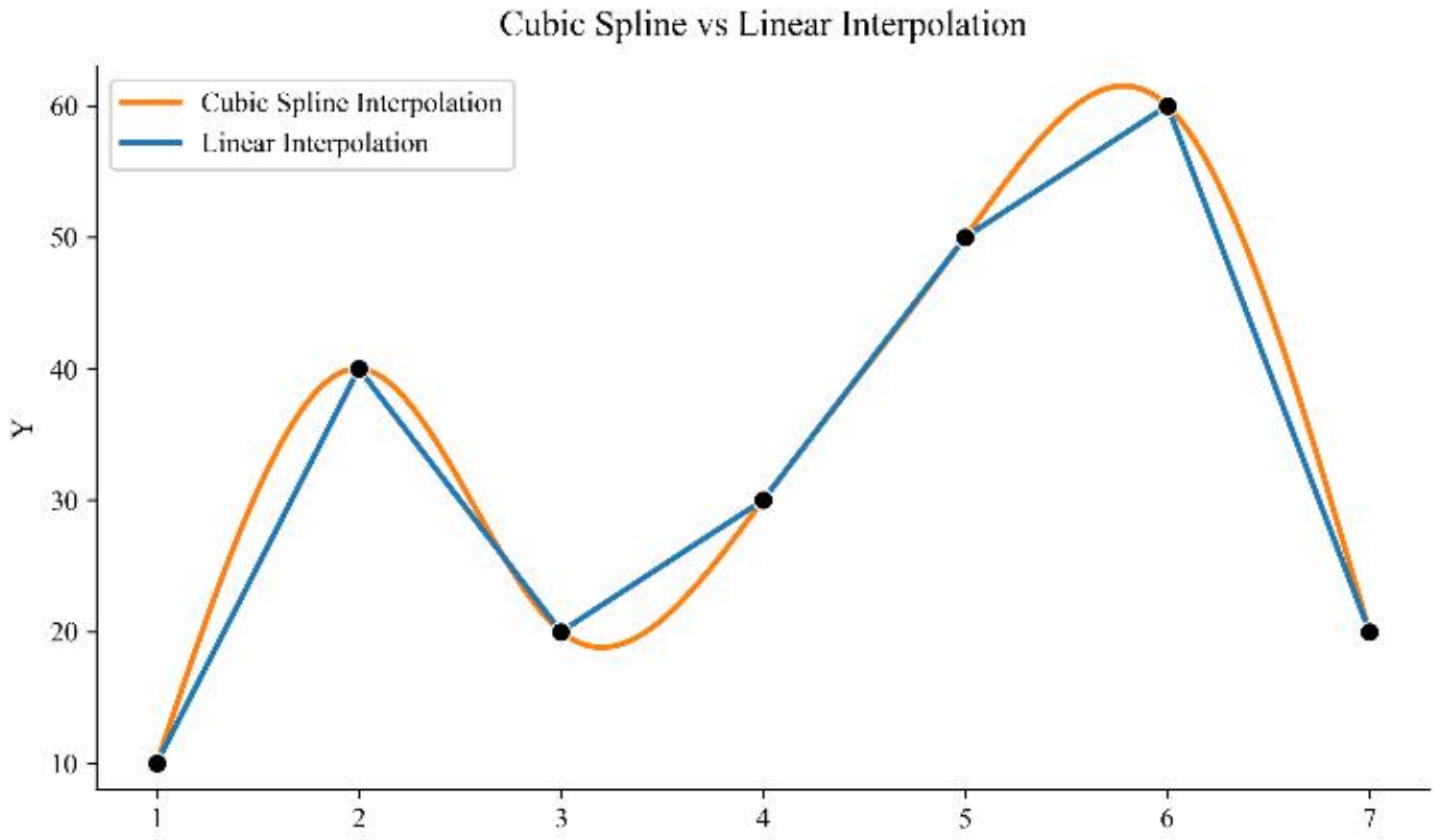

Fig. 4 Cubic Spline Interpolation Path Optimization Effect

### 3.2 Adaptive Affine Deformation Driven by Global Path

Upon obtaining the global smooth path $P_{TAPE}$, the leader centroid $p_{cm}$ moves point-by-point according to the path index. To achieve collision-free traversal through narrow corridors, the formation must deform during movement. Distinct from traditional conformal transformations that only permit uniform global scaling, the affine transformation matrix $A_c(t)$ introduced in this paper allows the formation to undergo non-uniform scaling and shearing. The affine transformation matrix $A_c(t)$ comprises a rotation matrix $A_{rot}(t)$ and a scaling matrix $A_{sc}(t)$:

$$A_c(t) = A_{rot}(t)A_{sc}(t) = \begin{bmatrix} cos(\theta(t)) & -sin(\theta(t)) \\ sin(\theta(t)) & cos(\theta(t)) \end{bmatrix} \begin{bmatrix} sc(t) & 0 \\ 0 & sc(t) \end{bmatrix} \tag{5}$$

The core parameters of the affine transformation matrix $A_c(t)$ are dynamically solved in real-time based on path deviation and obstacle distance:

Adaptive Adjustment of Rotation Angle $\theta(t)$: A directional error vector $p_{cms}$ is defined from the current position of the centroid to the target path point $P_{TAPE}(i_f)$. Based on the components of this vector along the X and Y axes, the algorithm calculates the required rotation angle of the formation in real-time, ensuring that the forward posture of the entire formation closely aligns with the tangential direction of the globally optimal trajectory.

Obstacle-Avoidance Adaptability of Scaling Coefficient $sc(t)$ : The crux of formation deformation lies in evading surrounding obstacles. The algorithm computes the shortest distance $|p_{cm} - q_{obsmin}|$ from the centroid $p_{cm}$ to the nearest obstacle boundary in real-time. Defining $kd_{max}$ as the safety distance threshold that triggers obstacle-avoidance deformation, the system perceives a collision risk when the obstacle distance is less than $kd_{max}$. At this point, the scaling coefficient $sc(t)$ is proportionally reduced based on the distance, driving the formation to shrink its shape to "squeeze" through the obstacle zone. When moving away from obstacles ( $|p_{cm} - q_{obsmin}| \geq kd_{max}$), the scaling coefficient is restored, and the formation automatically reconstructs its initial nominal dimensions，as shown in equation 6，$sc_{all}$ is the scaling coefffcient of the current formation relative to the nominal formation.

$$sc(t) = \begin{cases} \left(1 - \dfrac{kd_{max} - |p_{cm} - q_{obsmin}|}{kd_{max}}\right)\dfrac{1}{sc_{all}} & |p_{cm} - q_{obsmin}| > kd_{max} \\ \dfrac{1}{sc_{all}} & |p_{cm} - q_{obsmin}| < kd_{max} \end{cases} \tag{6}$$

By integrating the aforementioned parameters, the final real-time expected target position $pl_{tar}$ for the leaders is generated by the following control law:

$$\vec{p}_{tar} = P_{TAPE}(i_f) - p_{cm} \tag{7}$$

$$pl_{tar} = \left(\left(p_l + \vec{p}_{tar} - P_{TAPE}(i_f)\right) \times A_c(t)\right) + P_{TAPE}(i_f) \tag{8}$$

This formula perfectly actualizes the dual obstacle-avoidance mechanism of "macroscopically tracking the optimal path and microscopically executing elastic deformation."

### 3.3 Follower Consensus Tracking Control

After the leaders receive the affine deformation commands, the followers use a second-order integrator model for kinematic control. To eliminate the non-positive eigenvalues of the signed Laplacian matrix and ensure the stability of the system, a stable diagonal matrix $D = diag(d_i)$ is introduced [19]. Based on $D$, the original partitioned Laplacian matrices are transformed as follows:

$$\tilde{L}_{ff}^{s} = \left(D'' L_{ff}^{s}\right) \otimes I_d, \tilde{L}_{fl}^{s} = \left(D'' L_{fl}^{s}\right) \otimes I_d \tag{9}$$

where $D''$ is the corresponding submatrix of $D$, and $I_d$ is the identity matrix. Utilizing these transformed matrices, a distributed control law is constructed to drive the followers to track the leader formation. The consensus control protocol for the $i-th$ follower is designed as [19]:

$$u_i =- d_i \sum_{j \in N_i} w_{ij} \left[k_p \left(p_i - p_j + k_v(v_i - v_j)\right)\right] \tag{10}$$

Among them, $w_{ij}$ is a real-valued weight (which can be negative or positive) corresponding to the edge $(j, i)$, and $k_p$, $k_v$ are the positive control gains. Under this control protocol, the position tracking error $\delta_{pf}$ of the followers, i.e., the formation error [19], can be defined as follows:

$$\delta_{pf}(t) = p_f(t) - p_f^*(t) = p_f(t) + \tilde{L}_{ff}^{s^{-1}} \tilde{L}_{fl}^{s} p_l^*(t) \tag{11}$$

Macroscopically, this achieves a cooperative obstacle avoidance effect where the entire multi-UAV formation moves along the globally optimal path and locally deforms adaptively.

## 4. Experimental Design and Expected Results Analysis

To comprehensively evaluate the effectiveness and superiority of the proposed cooperative obstacle avoidance control system based on the BI-APF-RRT algorithm and affine transformation, multi-scenario comparative simulation experiments are conducted using the MATLAB platform.

### 4.1 Simulation Environment and Baseline Setup

The multi-agent formation in the experiments is configured to consist of seven unmanned aerial vehicles (UAVs), comprising three leaders and four followers. To ensure the fairness and rigor of the experiments, the underlying formation control in all comparative evaluations strictly adopts the identical "double-integrator follower control law and affine transformation mechanism." Consequently, the singular control variable across the tests is the global guiding path planning algorithm applied to the leader formation's centroid.

The following two representative algorithms are selected as baselines for comparative analysis:

Baseline A (Pure APF + Affine Control)[20]: Represents the prevailing mainstream methodology (i.e., the original scheme detailed in the literature), where the centroid's guiding trajectory is generated entirely by the traditional Artificial Potential Field (APF) method.

Baseline B (Traditional BI-RRT + Affine Control): Represents the fundamental sampling-based path planning approach, utilizing the conventional Bidirectional Rapidly-exploring Random Tree (BI-RRT) algorithm without potential field guidance to generate the centroid's path.

Proposed Algorithm (BI-APF-RRT + Affine Control): The advanced algorithm proposed in this paper, which uniquely integrates a goal-biasing strategy and a potential field stochastic perturbation mechanism.

Three highly challenging and representative obstacle scenarios are designed for the experiments. These scenarios are specifically formulated to rigorously evaluate the algorithms' capabilities in escaping local minima, maintaining the spatial coherence of affine deformations, and demonstrating comprehensive path optimization and tracking performance within high-density environments.

### 4.2 Local Minima Escape Experiment

When confronting "U-shaped" or concave obstacles, the traditional Artificial Potential Field (APF) method is highly susceptible to local minima deadlocks. This vulnerability arises because the attractive force exerted by the target and the repulsive force originating from the bottom of the obstacle become collinear and diametrically opposed, resulting in mutual cancellation. To rigorously evaluate the escape capabilities of the algorithms, this experiment staggers multiple miniature U-shaped traps in the center of the environment, with their openings directly facing the starting point, as illustrated in Figure [5].

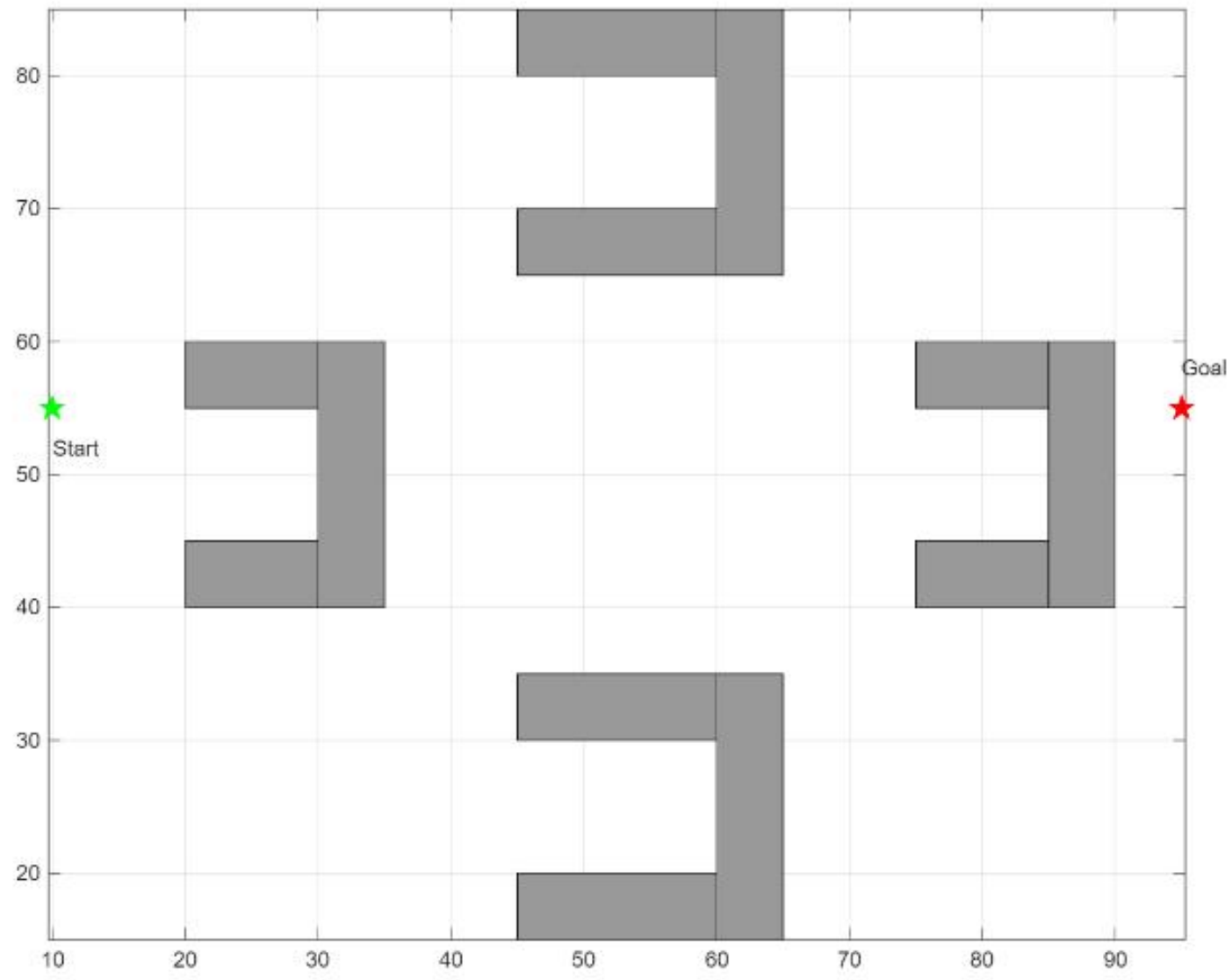

Fig. 5 U-Shaped/Concave Trap Environment

The simulation results demonstrate that when Baseline A (pure APF) guides the formation into the interior of the first U-shaped trap, the resultant force acting on the centroid converges to zero, causing the guiding trajectory to stagnate. Therefore, due to the lack of effective feedforward guidance, the underlying affine formation exhibits a standstill within narrow dead ends, as shown in Figure 6. It can also be seen from Figure 7 that the metric remains unchanged after a certain period of time. Ultimately, the formation fails to navigate out of the trap, yielding an obstacle avoidance success rate of 0%.

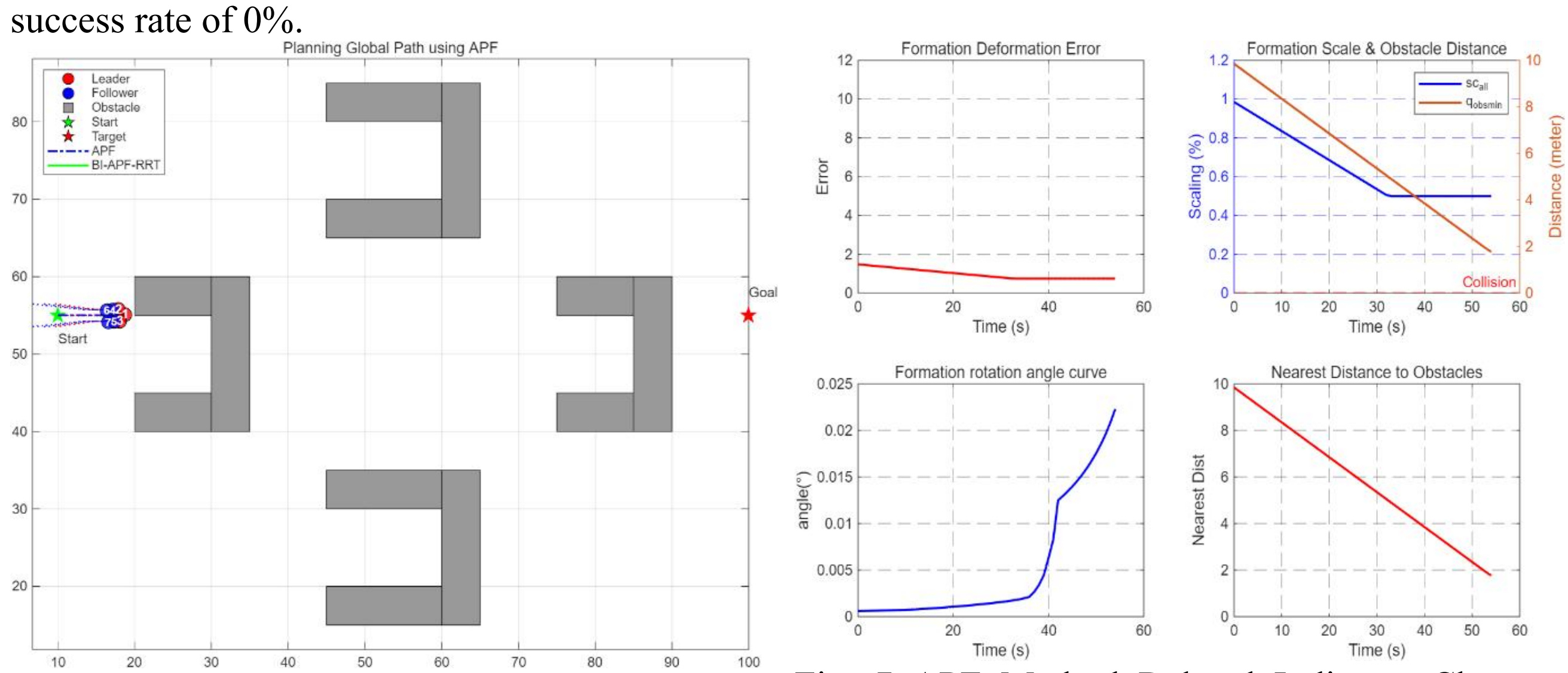

Fig. 6 APF-Based Affine Transformation Adaptive Obstacle Avoidance （U-Shaped/Concave Trap Environment）

Fig. 7 APF Method Related Indicator Change Curve（$\delta_{pf}$、$sc$、$\theta$、$nearest\ dist\ to\ obs$）

In stark contrast, the BI-APF-RRT algorithm proposed in this paper successfully overcomes the fatal flaws inherent in traditional methods when navigating dead-end environments. By leveraging the introduced stochastic perturbation mechanism and goal-biasing strategy, the algorithm macroscopically perceives the spatial distribution of the traps and decisively delineates a smooth, "S"-shaped traversing detour trajectory. Driven by this optimal guiding path, the underlying affine formation seamlessly scales down and maneuvers laterally past the obstacle boundaries. The entire obstacle avoidance execution is remarkably fluid, as shown in Figure 8, devoid of any deadlocks or collisions, thereby achieving a 100% task completion rate. Figure 9 shows the changes in formation tracking error, scaling factor, rotation angle, and the nearest distance to obstacles during this process.

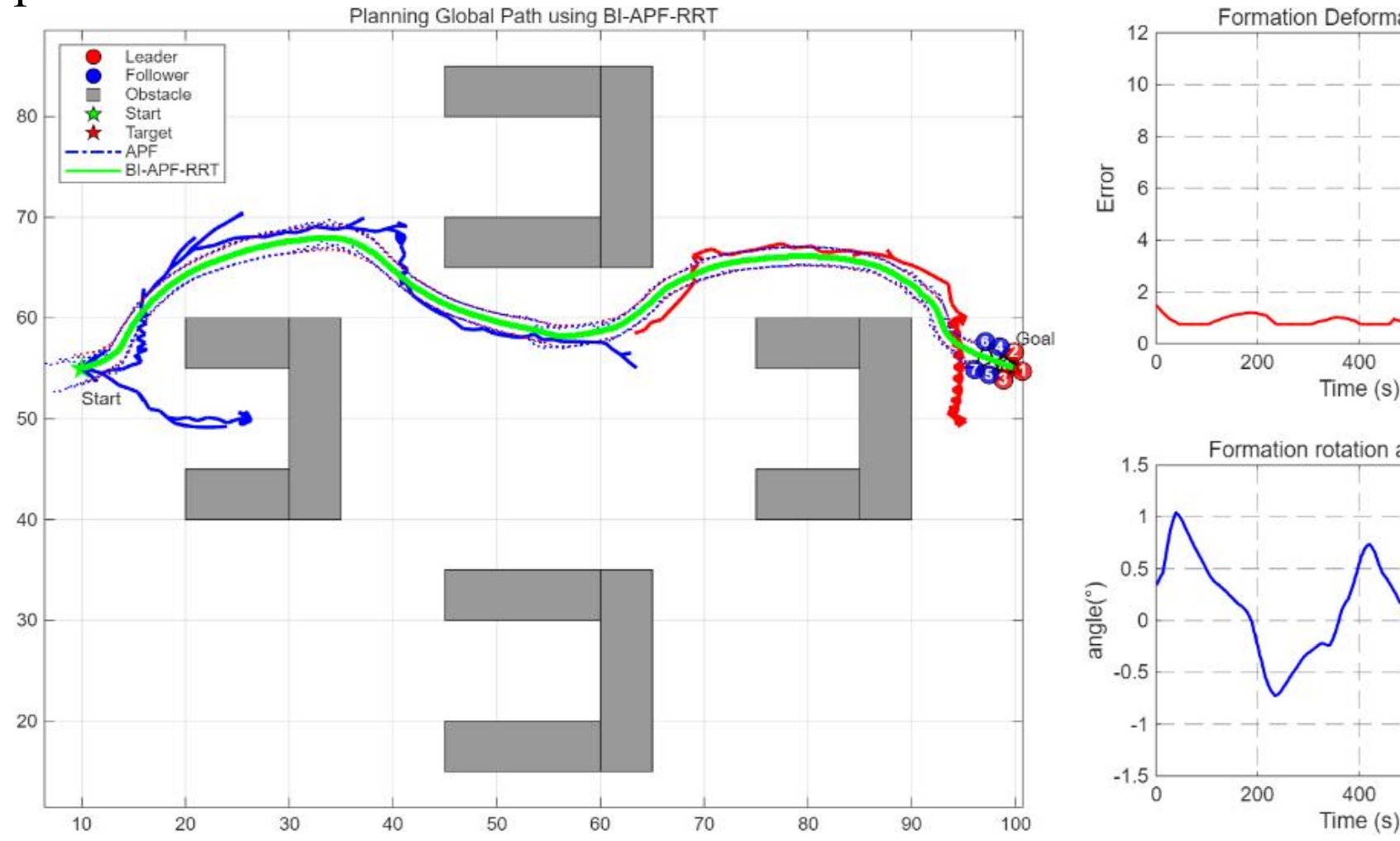

Fig. 8 BI-APF-RRT-Based Affine Transformation Adaptive Obstacle Avoidance（U-Shaped/Concave Trap Environment）

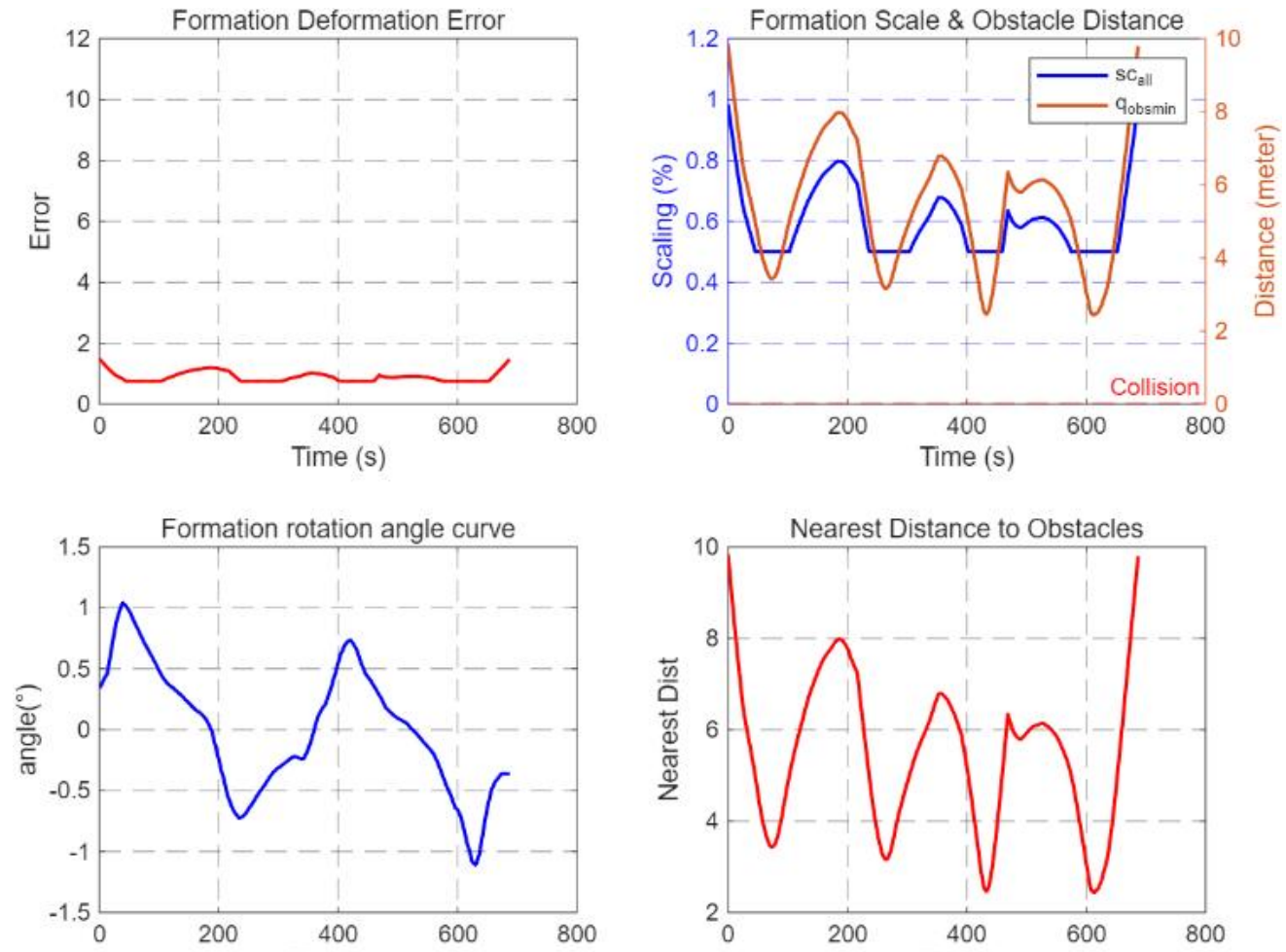

Fig. 9 BI-APF-RRT Method Related Indicator Change Curve（$\delta_{pf}$、$sc$、$\theta$、$nearest\ dist\ to\ obs$）

### 4.3 Complex Asymmetric Narrow Channel Experiment

In order to test the elastic adaptability of the formation in extremely constrained spaces, this experiment designed a continuous 'stair-step diagonal narrow corridor,' as shown in the figure 10. The junctions of this corridor are extremely narrow, and the overall layout is diagonally distributed, which requires the formation to undergo both deep 'non-uniform scaling' and 'continuous rotation' simultaneously in order to pass safely.

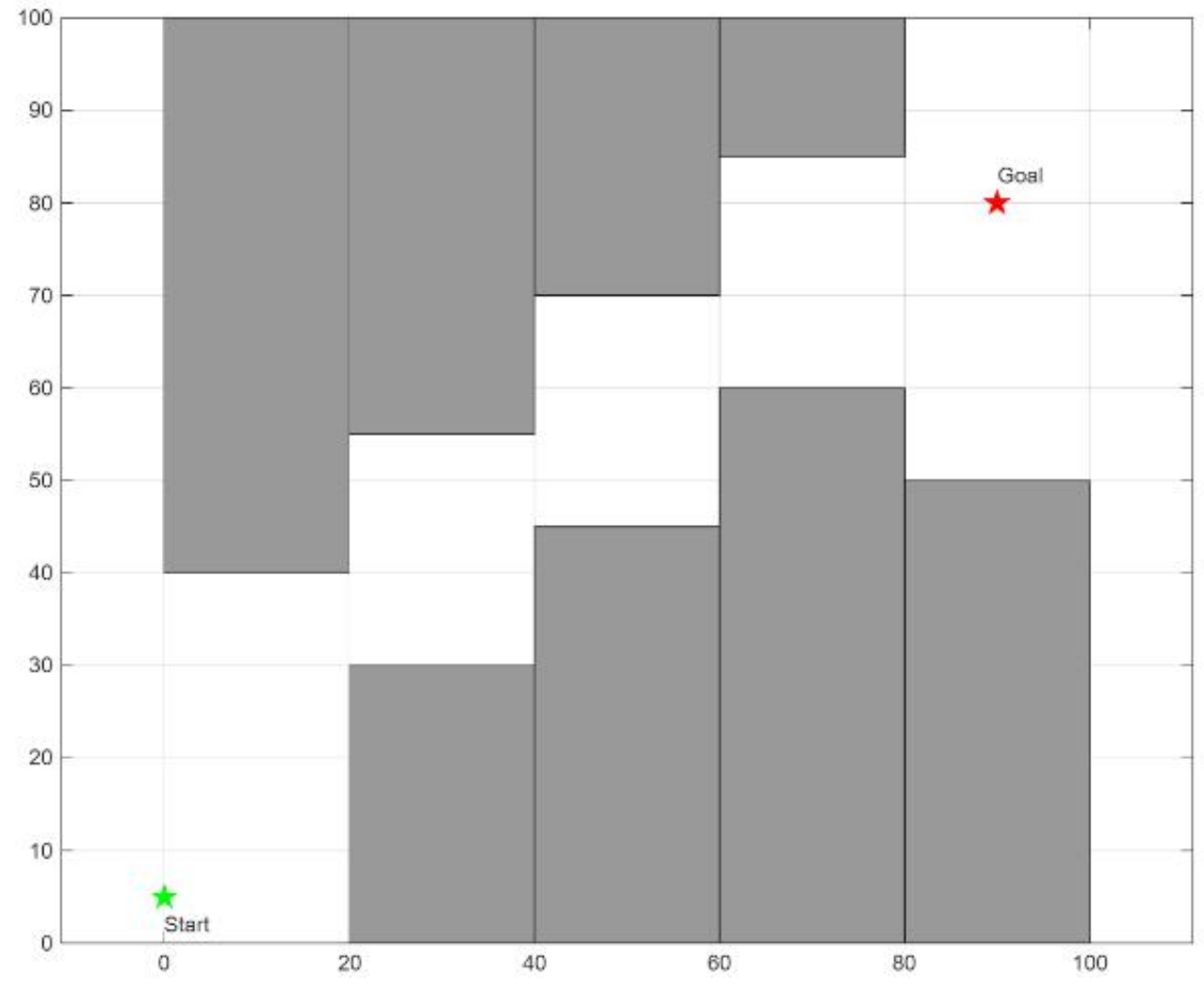

Fig. 10 Continuous Stair-step Diagonal Narrow Corridor Environment

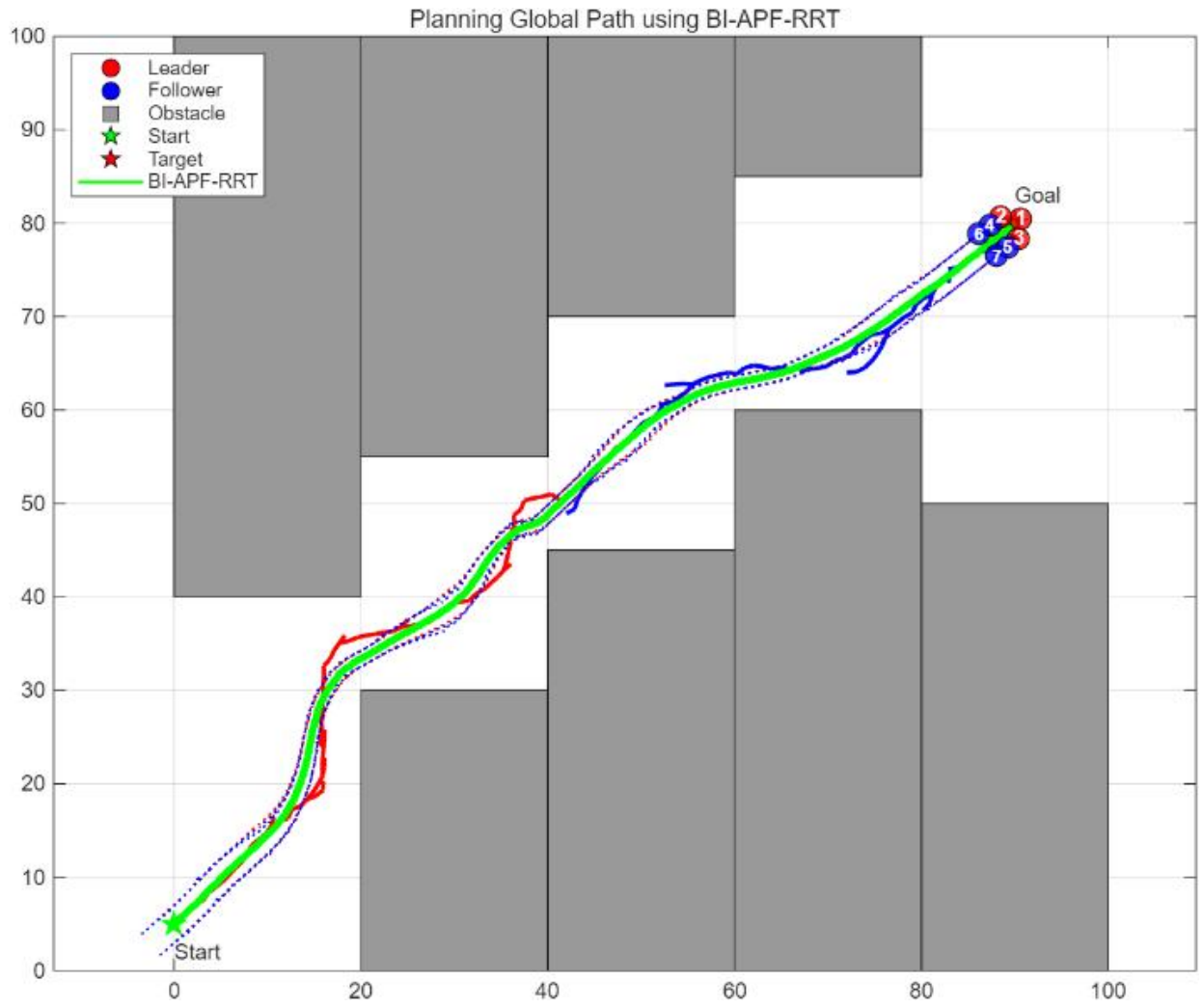

Fig. 11 Path Planning Based on BI-APF-RRT Algorithm (Continuous Stair-step Diagonal Narrow Corridor Environment)

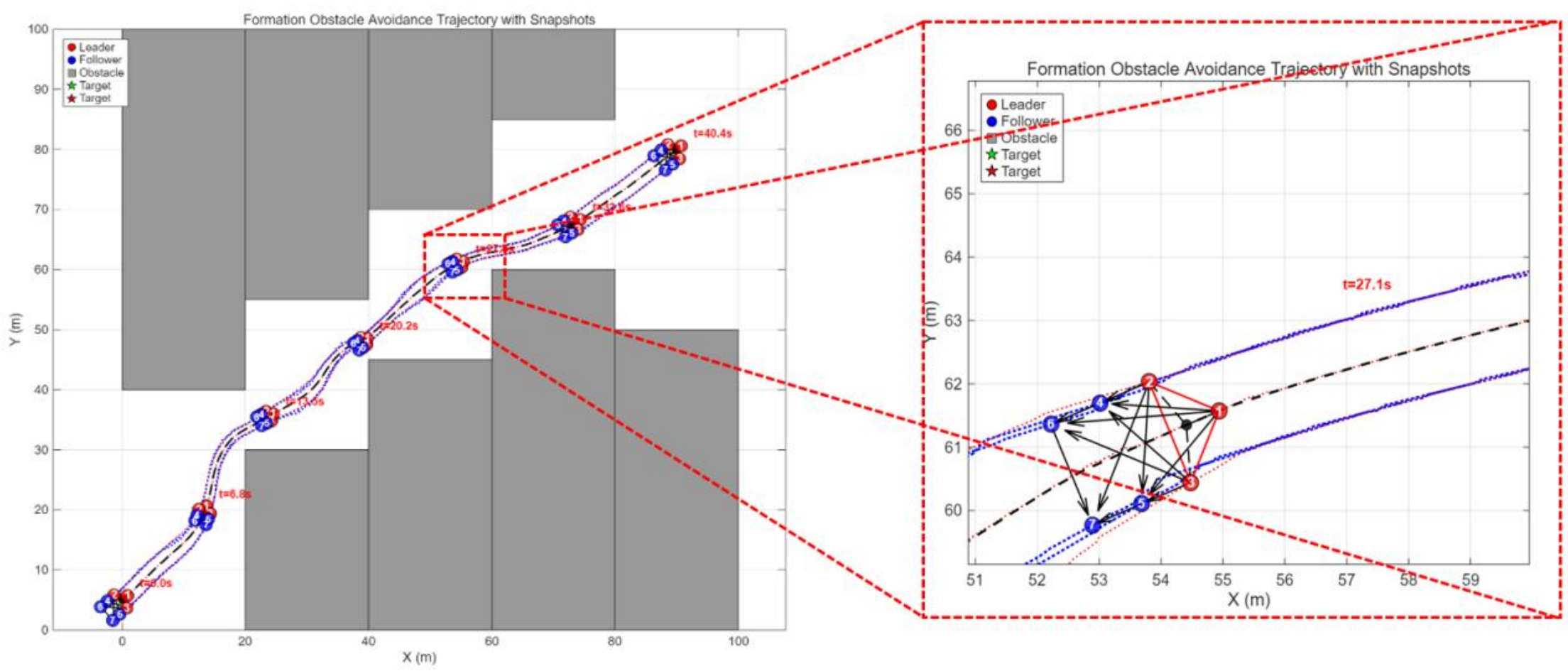

Fig. 12 Formation Snapshot and Formation Operation Trajectory

Figure 11 shows the BI-APF-RRT algorithm planning a path in a continuous stair-like oblique narrow corridor environment, and Figure 12 shows the UAV formation at different time periods, as well as the trajectories of each UAV.

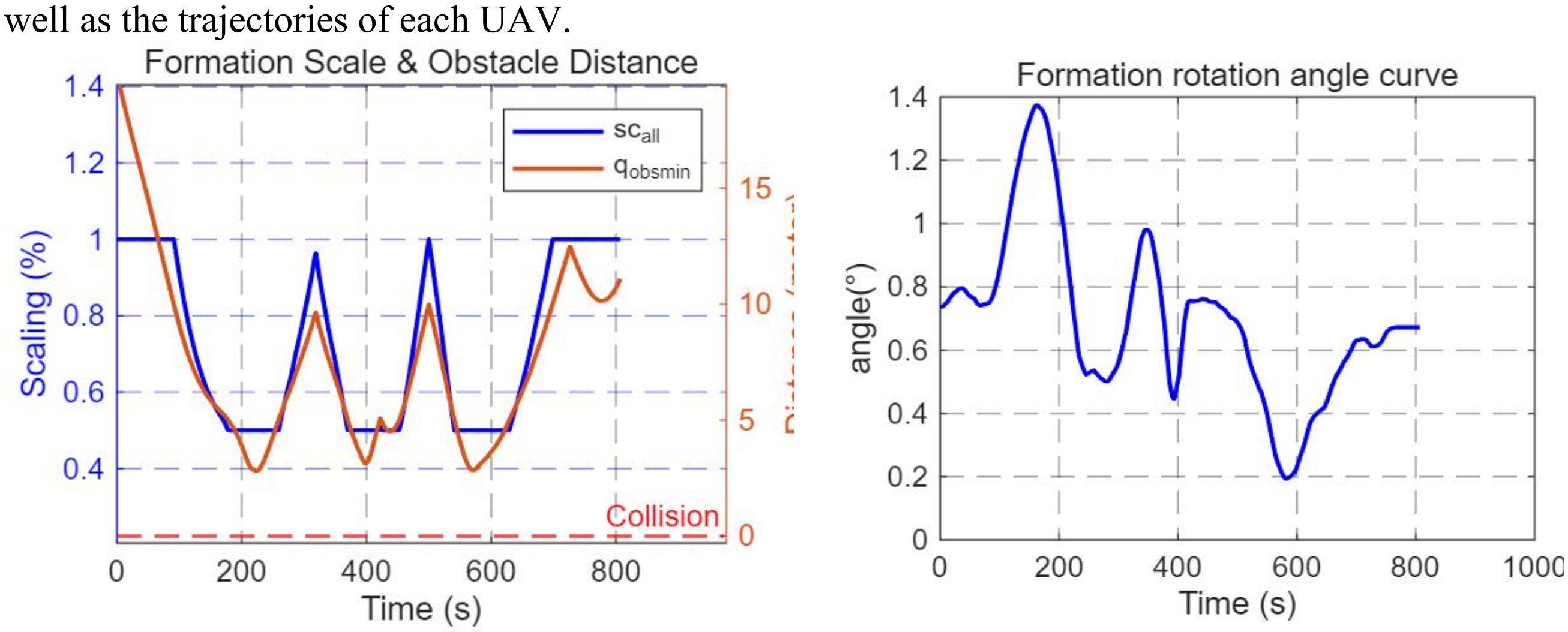

Fig. 13 Change Curve of The Scaling Factor $sc$ in The Continuous Stair-step Diagonal Narrow Corridor Environment (BI-APF-RRT Method)

Fig. 14 Change Curve of The Rotation Angle $\theta$ in The Continuous Stair-step Diagonal Narrow Corridor Environment (BI-APF-RRT

Method)

It can be clearly observed through Figures 13 and 14:

The precise contraction and recovery of the scaling factor $sc(t)$: When the formation's centroid approaches the edge of an extremely narrow corner of the passage (with obstacle distance sharply decreasing), the $sc(t)$ curve rapidly drops, and the underlying follower network is instantaneously 'flattened' to adapt to the narrow gap; after leaving the tight area, $sc(t)$ smoothly recovers to 100%, and the formation automatically reconstructs to its nominal size.

Adaptive alignment of the rotation angle $\theta(t)$: When facing a staircase-like diagonal passage, the $\theta(t)$ curve shows continuous and smooth deflection responses. This demonstrates that the formation is actively adjusting its posture to conform to the diagonal guidance line planned by BI-APF-RRT with the narrowest side profile. This experiment fully confirms that, thanks to the smooth trajectory provided by BI-APF-RRT, the formation can adaptively adjust affine matrix parameters like flowing water, completely demonstrating excellent compliant obstacle avoidance maneuverability within constrained spaces.

### 4.4 Comprehensive Performance Evaluation of High-Density Obstacle Environments

In order to comprehensively quantify and evaluate the overall performance of the algorithm, in this experiment, 18 complex obstacles of varying sizes and positions were randomly distributed within a 100 × 100 space, and 20 independent Monte Carlo tests were conducted. This section mainly involves a comparative analysis with Algorithm B (the traditional BI-RRT).

1. Comparison of macro-level planning efficiency

Table 1 lists the average planning metrics of the two algorithms over 20 tests. The data indicate that, due to the traditional BI-RRT using blind global random sampling, the search tree diverges excessively in space; whereas the APF potential field introduced in BI-APF-RRT greatly enhances the directionality of the search. Compared with Algorithm B, the BI-APF-RRT algorithm in this paper reduced the average number of iterations by about 25.6%, shortened the average path length by 13.7%, and reduced the average planning time by about 32.3%, demonstrating extremely high online planning efficiency.

Table1 Algorithm Planning Performance Comparison in High-Density Random Environments (Average of 20 Runs)

| Algorithm | Average planning time (s) | Average number of iterations | Average path length(m) |
|---|---|---|---|
| BI-RRT | 0.68 | 956 | 174 |
| BI-APF-RRT | 0.46 | 711 | 150 |

2. Microscopic Tracking Error and Dynamic Stability Analysis

Figures 15 and 16 depicts the comparison curves of the formation tracking errors of the lower-level followers when performing obstacle avoidance deformation in the same high-density map.

The comparative error curves clearly illustrate the dynamic limitations of the traditional BI-RRT method. Due to its convoluted paths and non-smooth turning points, the underlying second-order integrator control law exhibits significant tracking deviations, resulting in abrupt, impulse-like spikes (peaking exactly at 2.0) during obstacle avoidance maneuvers.

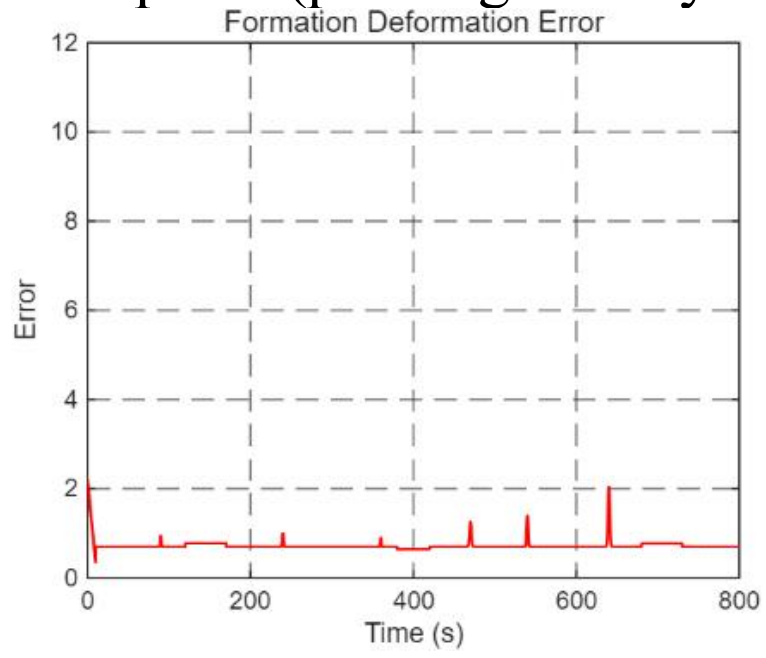

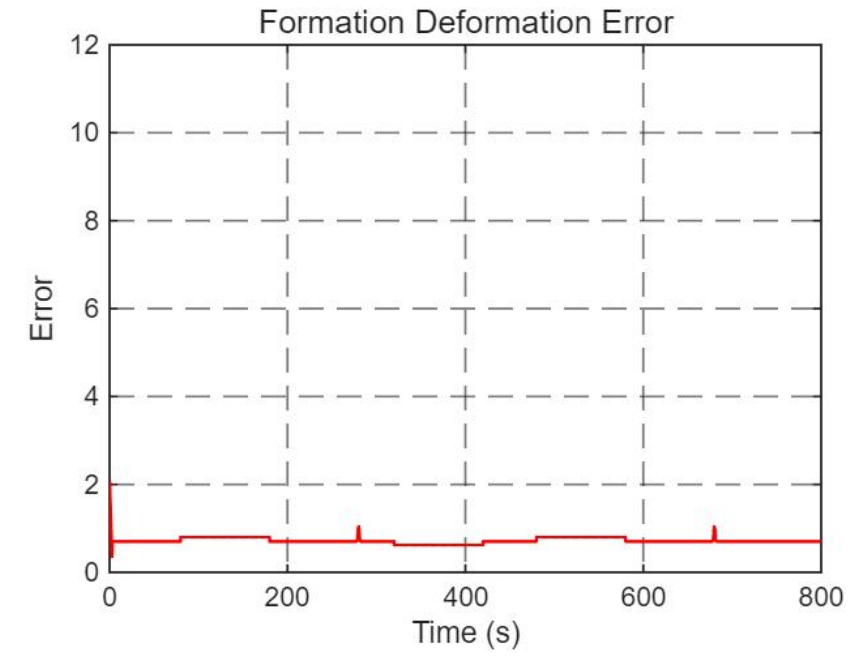

Fig. 15 Tracking error curve（BI-RRT） Fig. 16 Tracking error curve（BI-APF-RRT）

Conversely, the proposed BI-APF-RRT leverages potential field guidance and cubic B-spline interpolation to ensure geometric path continuity. As shown by the error curve, the system quickly converges from its initial state to a low, stable baseline error. During complex shuttle operations involving frequent formation compression and rotation for obstacle avoidance, the deformation error experiences only minor, strictly bounded fluctuations (peaking merely around 1.0), successfully eliminating the severe transient spikes observed in the traditional method. Furthermore, it swiftly recovers to the baseline post-maneuver. This confirms that our highly smooth global guidance path effectively minimizes the dynamic tracking burden on the formation, securing both physical connectivity and structural stability in highly constrained environments.

## 5. Conclusion

Aiming at the local minimum defects and formation structure maintenance difficulties in multi-UAV formation obstacle avoidance, this paper innovatively proposes a cooperative obstacle avoidance algorithm combining BI-APF-RRT global path planning with real-time affine transformations. By using a bidirectional RRT algorithm with goal-biasing and B-spline smoothing at the centroid guidance level, the global optimality and dynamic feasibility of the guiding trajectory in complex obstacle environments are ensured. Concurrently, by establishing an "obstacle avoidance distance - affine scaling" feedback mechanism, the elastic deformation of the entire formation when passing through narrow areas is achieved. This algorithm effectively balances the flexibility of obstacle avoidance with the integrity of the formation structure, providing a reliable theoretical basis and solution for the practical deployment of multi-agent swarms in complex confined spaces.

**Funding**： No funding was received for this study.

**Competing Interests**： The authors declare that they have no competing interests to disclose.